# Extracting Region of Interest for Palm Print Authentication

**Mrs. Kasturika B. Ray**
**Dean (*Academics*)**
**Eklavya College of Engineering Technology and Science**
**Kusumati,Bhubaneswar,Odisha,India**

**Institute of Technical Education & Research.**

**Siksha 'O' Anusandhan University**

**Bhubaneswar (Phd contd.)**

**E-mail** – **kasturikabray@rediffmail.com**

**Mrs. Rachita Misra**
**Prof. & Head Department of IT**
**CV Raman College of Engineering**
**Bhubaneswar, Orissa, India.**
**E – mail – rachita03@yahoo.com**

Kasturika B. Ray

**Abstract-** Biometrics authentication is an effective method for automatically recognizing individuals. The authentication consists of an enrollment phase and an identification or verification phase. In the stages of enrollment known (training) samples after the pre-processing stage are used for suitable feature extraction to generate the template database. In the verification stage, the test sample is similarly pre processed and subjected to feature extraction modules, and then it is matched with the training feature templates to decide whether it is a genuine or not.

This paper presents use of a region of interest (ROI) for palm print technology. First some of the existing methods for palm print identification have been introduced. Then focus has been given on extraction of a suitable smaller region from the acquired palm print to improve the identification method accuracy. Several existing work in the topic of region extraction have been examined. Subsequently, a simple and original method has then proposed for locating the ROI that can be effectively used for palm print analysis. The ROI extracted using this new technique is suitable for different types of processing as it creates a rectangular or square area around the center of activity represented by the lines, wrinkles and ridges of the palm print.

The effectiveness of the ROI approach has been tested by integrating it with a texture based identification / authentication system proposed earlier. The improvement has been shown by comparing the identification accuracy rate before and after the ROI pre-processing.

## 1. Introduction

With the development advanced systems which provide service after checking the identity of a person, improving the personal identification methods have become a necessaity. In these identification techniques, biometric features are extracted from human biological organs or behavior.

It can be considered as the technology that describes procedure for identification and verification using techniques of extraction, storage and matching of features from digitized biometric images such as Finger Print, Face, Iris and Palm Print. The main characteristics of palm print images are basically of three types: the principal lines, smaller lines or wrinkles and ridges, which have been used to extract features by Zhang et al [1]. Various biometric technologies for personal identification have been presented by Jain et al. [2]. Several researchers have acquired image features, classified and identified using different methods [3, 4, 5, 6]. In general palm print identification methods available in the literature can be classified as Geometric, Statistical and Textural feature extraction methods [11].

A central part of palm print identification systems is similarity comparison between feature vectors of test sample and known templates, and improve the accuracy of correct detection or authentication. In general, the comparison is performed by using various distance measures. There have been considerable efforts in the palm print biometric literature for determining a suitable region (referred as ROI) before applying the feature extraction methods.

Use of Sobel and Morphological operations of palm print was found to be suitable in many network-based applications analyzed by Chin chuan Han et al [4]. In this paper they have suggested region extraction steps to obtain a square region in a palm table which is called region of interest (ROI). A method of locating and segmenting the palm print into region of interest (ROI) using elliptical half-rings has been reported by Poon et al [5] to improve the identification.

The use of ROI while applying correlation filter classifiers for palm print identification and verification has been reported by Pablo Hennings [6]. In this work three different regions of pixel sizes: 64×64, 96×96, 128×128 has been used. Two reference points were first determined from the hand geometry, and square regions are extracted after aligning these two points with the vertical axis.

J.Z. Wang, J. Li, and G. Wiederhold have proposed an integrated region matching (IRM) scheme which allows for matching a region of one image to several regions of another image and thus decreases the impact of inaccurate segmentation by smoothing over the imprecision. The scheme is implemented as SIMPLIcity system [7].

Ying-Han Pang et al have used various moments (ZM, PZM and LM) as feature descriptors [8]. In the first stage of their experiment a localization of palm print region has been implemented as per methodology given by Tee Connie et al [9]. Different methods in Palm print Feature extraction using Region-of-Interest has been analyzed by Kasturika et al [10].

A region based recognition system imitates the human cognition. The objective is to eliminate image areas that do not contribute to features that are useful in differentiating or classifying images. It tries to redefine the boundary within the acquired image. This smaller region becomes the focus area for applying various feature extraction and classification strategies. It is expected that the region of interest will improve the accuracy of identification and authentication.

This paper presents an effort to extract a rectangular region from the acquired image using simple operations. The effectivenes of the ROI extraction method has been examined by comparing identification accuracy before and after using the ROI.

## 2. Palm print Recognition Processing

A palm print verification system can be thought of as five modules namely, palm print acquisition, preprocessing, feature extraction, storage and feature matching. The authentication system consists of two stages, first enrollment of known samples and next assignment or verification of unknown samples. In the enrollment stage, the training samples are collected from the pre-processing stage, features are extracted and it goes to the storage as templates. In the verification stage, a test sample is processed by the preprocessing and feature extraction modules, and the extracted feature set is matched with the training samples in the template feature database to decide whether it is a genuine or not. An identification system classifies the test sample as one of the training sample based on a similarity measure.

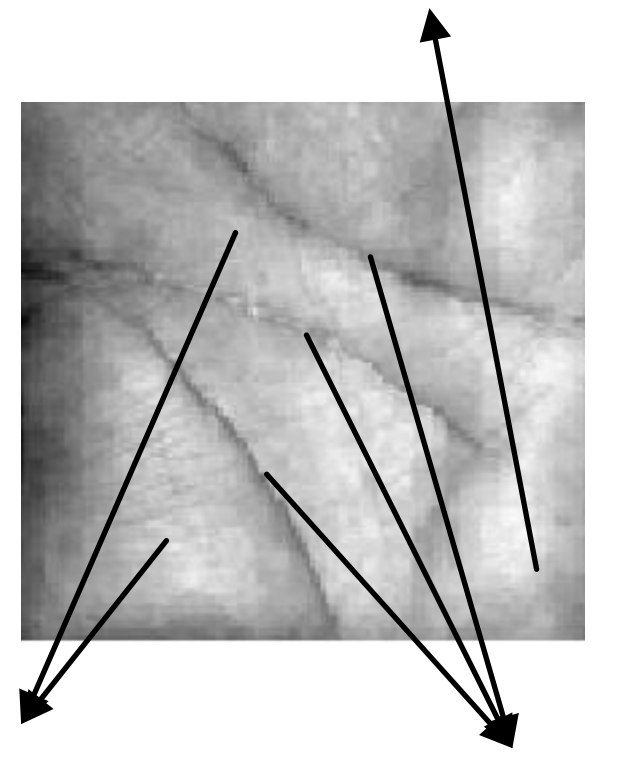

**Fig. 1: Palm-print image with principal lines, wrinkles and ridges**

The visible characteristics of the palm print are the Principal lines, Wrinkles and Ridges (Fig 1). While hand geometry focuses on principal lines statistical and use of transforms consider the general placement of these visible features. However, wrinkles play an important role in palm print authentication. Textural methods can be thought of as methods that focus on the detail pattern giving importance to wrinkles and ridges. So texture analysis can be suitably applied for palm print authentication.

A textural approach using edge gradients for palm print feature extraction and identification was reported earlier [12]. In this approach textural characteristics of different sub regions of the palm had been used as the feature vector. The “busyness” or “edgyness” of the sub-region represented the textural characteristic of the inherent wrinkles, ridges and principal lines. A new method of extracting a Region of Interest in the Pre-processing stage which improves the rate of correct identification has been described in the next section.

## 3. Region Extraction Approach

The new approach to ROI extraction is based on the principel of selecting a region where rich texture patterns can be found. In other words we need to discard those parts of the acquired palm print image which do not contribute to textural features, or are entirely different from the main region texture. Some initial experiments with different palms provides a clue that we need to find the central location of “busyness” and “edgyness” of each palm and construct a suitable ROI around this location. Our aim is to use a method which will provide us a square or a rectangular ROI with maximum number of pixels.

The acquired image has been divided into smaller horizontal and vertical strips and the statistical properties of the edgyness of these regions have been used to either select or reject the strips from the ROI. A comparison of Histogram of the original and extracted image is used to confirm the suitability of extracted region. Histogram techniques are used to check suitability of extracted region. The detailed steps for ROI extraction is illustrated in the next section.

This method of ROI extraction can be used as a preprocessing step for any of the palm print identification strategies in the literature. To verify its suitability the preprocessing has been applied to the identification method that was reported earlier [12].

## 4. Region Extraction Steps

To carry out the experiment, palm print images have been taken from internet (total 600 peg free poly u database collected from Hong Kong polytechnic university). The database consists of palm print of 60 different individuals, each having 12 samples of left palm and 12 samples of right palms. In these images, the entire palm is preserved with fingers and thumb being omitted. The image size is 384x284 pixels in 256 gray levels.

As the first step the original picture of size 284 X 384 pixels has been divided into vertical strips of 10 pixels each. The strips obtained from left side can be named as '*lea','leb',...,'leal' corresponding to the 38 sets of vertical lines 1-10,11-20…….375-380*. These strips can also be named as '*ra', 'rb', ... 'ral'* from right to left (see Fig 2 (a) and (b)). Next the picture is divided into horizontal strips of 10 pixels each, thereby giving 28 strips. The strips obtained from upper part to lower part of the figure is denoted as *'ua', 'ub', .. 'uab'*. Similarly the strips from lower to upper part f the picture is denoted as '*loa', 'lob', ..., loab'* (see Fig 2 (c) and (d)).

After this subdivision process ( as illustrated in the figures 2), the 'busyness' of these strips are determined by counting the number of connected lines (Numlines). The Sobel operator with 8 pixel window has been used to detect connected lines. Now considering the horizontal strip, the average Numlines and the standard deviation is determined. A threshold 'T' is then computed as;

$$T = mean - n * stddev,$$

Where, n is chosen as 1.The purpose of this threshold is to select or reject a strip as part of the ROI for a picture. Accordingly thresholds $T_h$ and $T_v$ for the horizontal and vertical strips are fixed. A ROI for this picture is then obtained by using this threshold on the horizontal strips as well as the vertical strips. As can be seen from Fig 2 e to f, some of the end strips are rejected as the 'busyness' in these areas are lower than the central part of the palm (lower than threshold). This technique is able to acieve a ROI in the central region of the picture with respect to the palm features such as principal lines, wrinkles and ridges.

When this concept is directly applied over the set of training and test palm images the ROIs obtained will be of unequal size. Also the boundaries of the strips on left, right, top and bottom will be different. To get a uniform size ROI which is simillarly located within all palms, we can use the intersection of these ROIs. This logical, however, this will give a very small area, which may eliminate some useful features. Instead a measure such as the number of times (frequency) a strip is within the ROI across the set of palm prints can be considered. A maximum frequency criterion is therefore used to fix the left, right, and top and bottom boundaries. Such a criterion will be flexible and adaptive over different images.

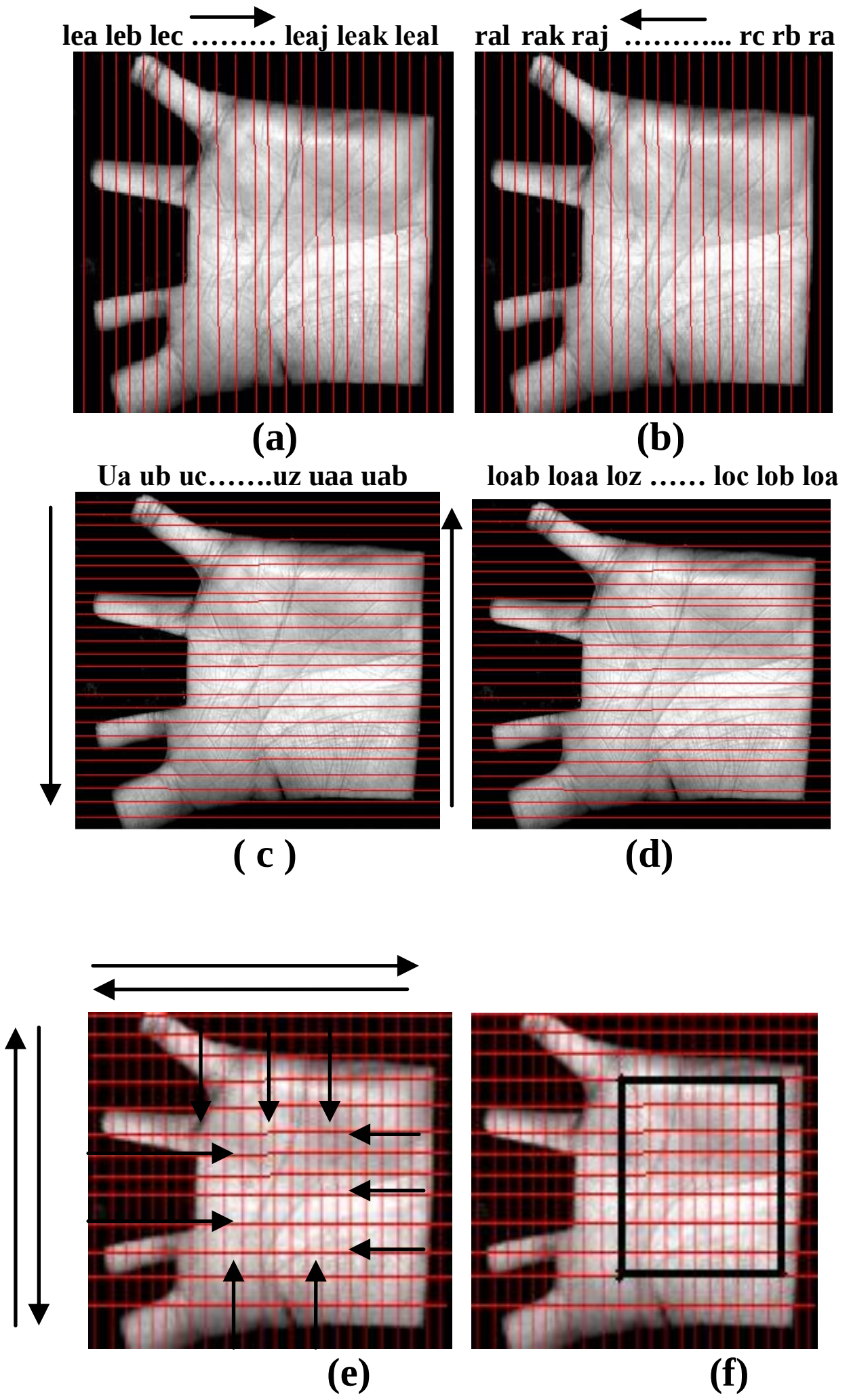

**Fig. 2 : Representation of Palm image coordinate diagram (a) key points of left to right (b) key points of right to left ( c ) key points of upper to lower (d) key points of lower to upper (e) discard the edgyness region from four sides (f) selected central region of a palm .**

The ROI obtained using this new technique has been compared with the original picture for conservation of palm details, by using Histogram comparison. As can be seen in Fig 3, the peak of the histograms of original palm and ROI are approximately same. In some cases the original histogram is bi-modal, where as the ROI is uni-modal. This gives a confidence that the process is able to eliminate some of the unnecessary part of the original image while preserving the important features.

## 5. Palm print authentication using the ROI method

To verify palm print authentication using the ROI as suggested here, the texture methods of authentication that has been reported [12] earlier is taken. The selected sets of feature are now extracted over the ROI and can be stored for several palms (60 palms from 10 individuals). This forms the database of known palm prints stored as classified training feature database *(T)*. Figure 4 shows a few original training palms and the extracted ROI containing all the Principal lines forming the central part of palm features.

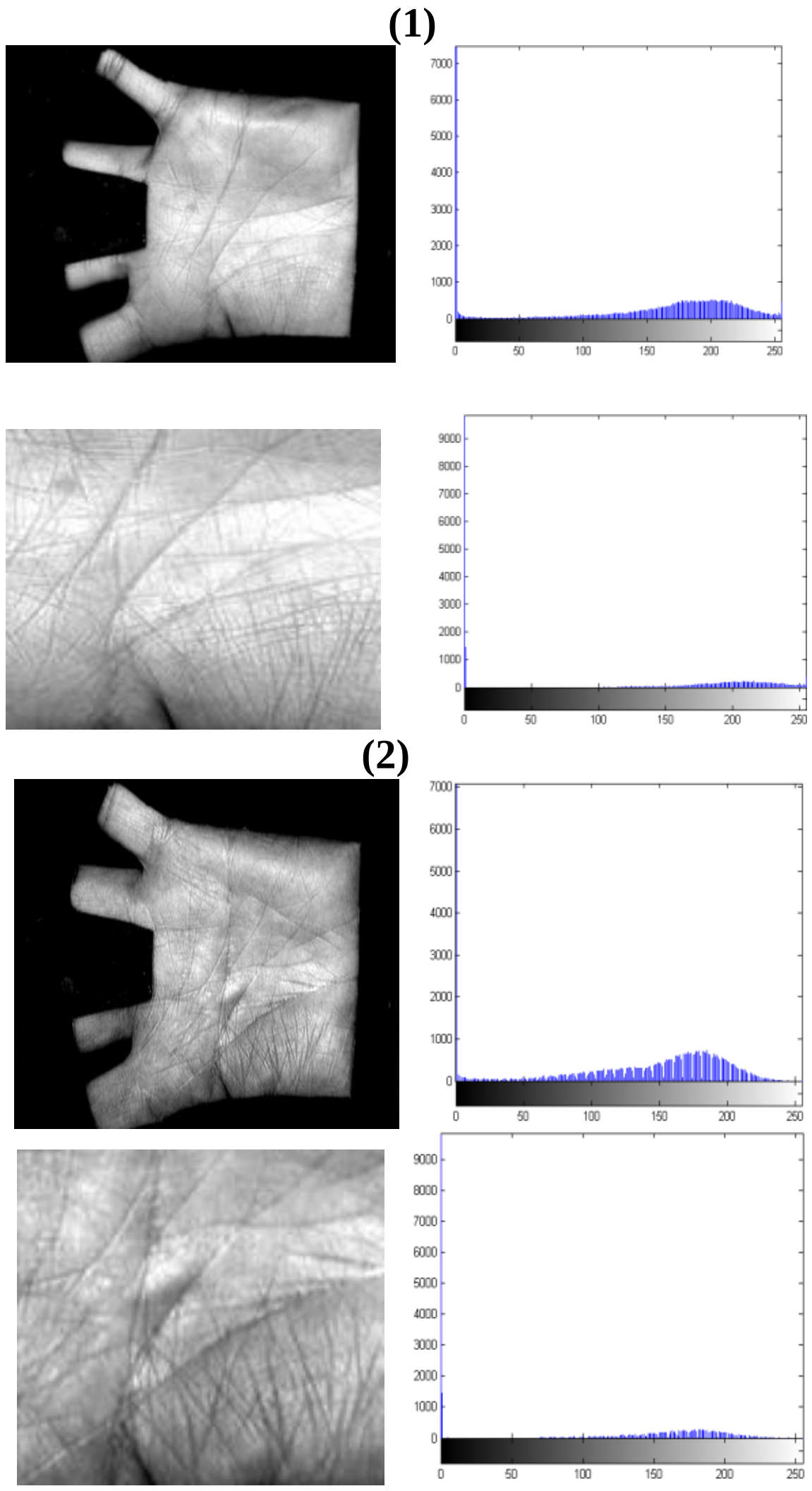

**(3)**

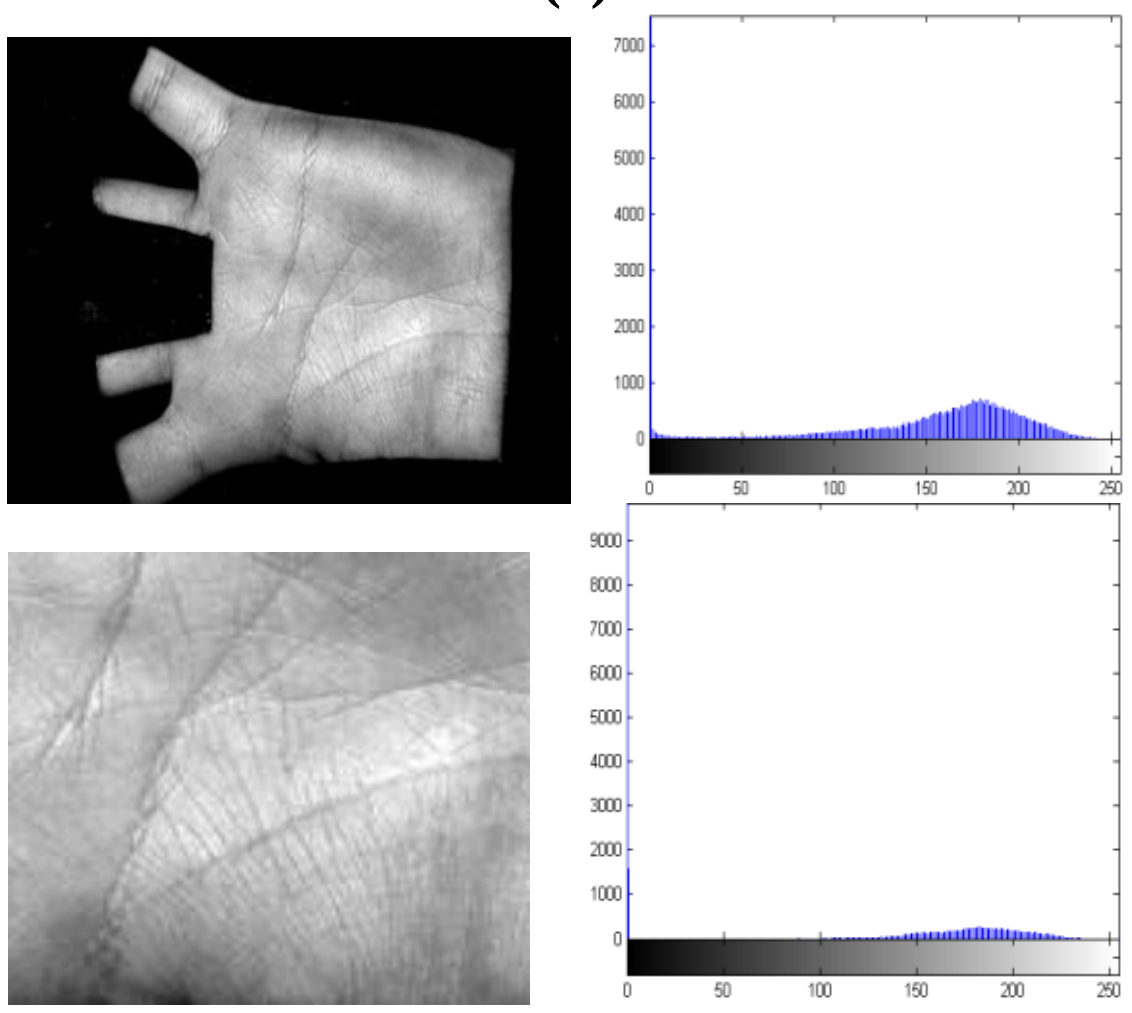

**(4)**

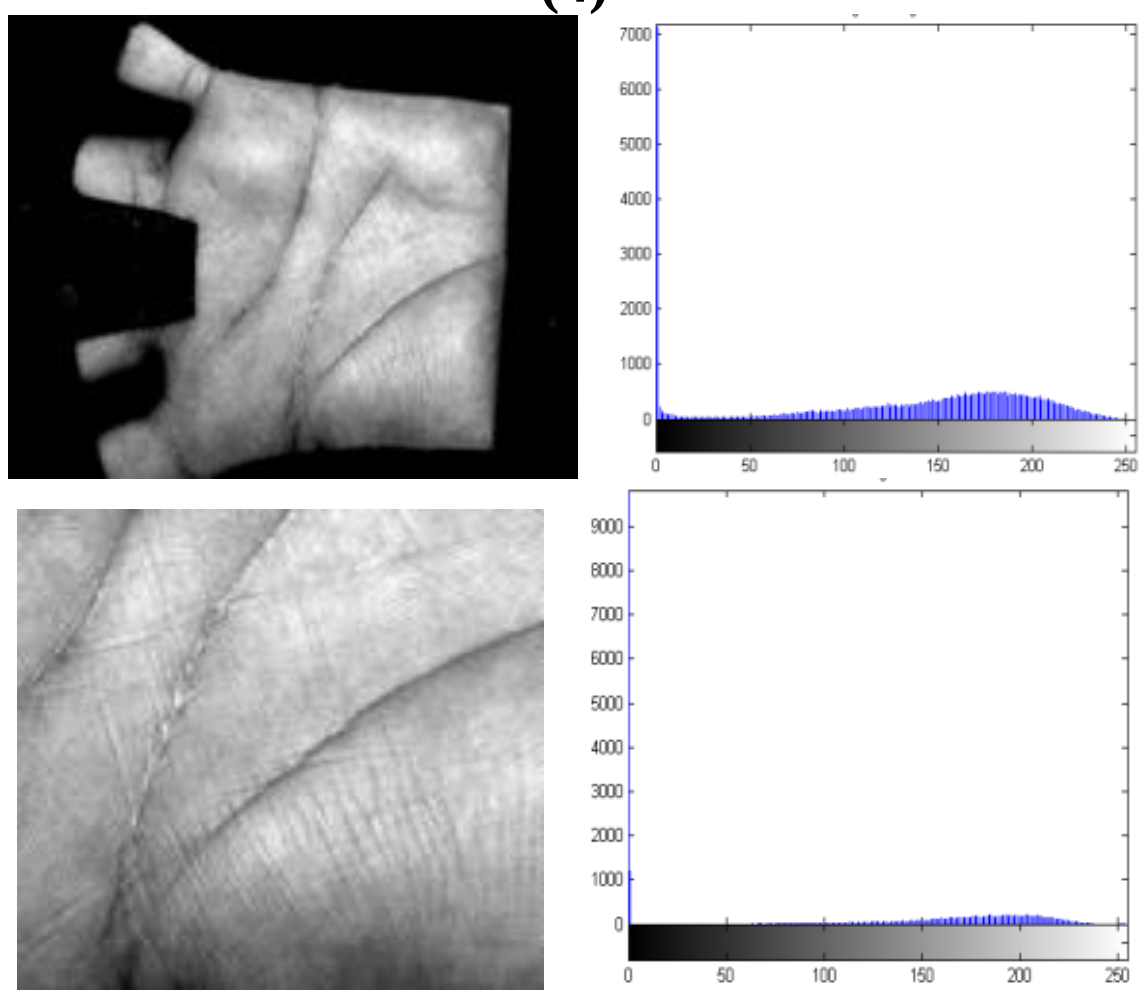

**Fig. 3: Four sample images of original palm, and generated ROI of those images with their histograms.**

Then the test samples are selected randomly from 60 palms of the test set (60 palms from 10 individuals). The similar texture features are extracted on the ROI of the test samples. The effectiveness of the authentication method using ROI was then examined based on the correct identification rate *(R)*, defined as:

$$R = \frac{\textit{No of test samples correctly classified}}{\textit{Total number of test samples selected}}$$

The accuracy rate R for feature sets 4, 8, and 16 before and after applying the ROI are calibrated in Table 1. A100% detection in rate for any feature vector size indicates that by using ROI preprocessing the feature vector length can be reduced.

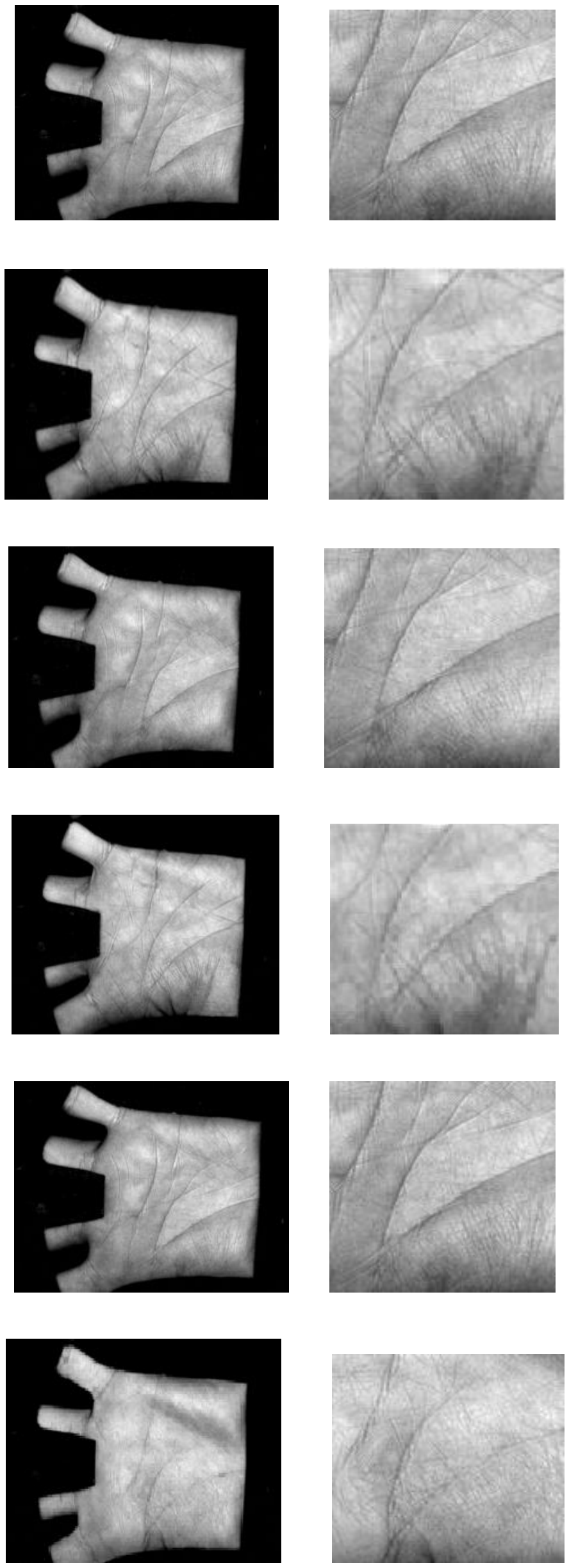

**Fig. 4 : Six Input palm print images for preprocessing and their generated central parts of the image as ROI**

**Table 1. Comparison of Performance between previous work and current work.**

| Image / ROI Size | Feature vector size | Accuracy of detection |
|---|---|---|
| Full palm print , 384 x 284 pixels in 256 gray levels | 4 | 70% |
| | 8 | 80% |
| | 16 | 90% |
| ROI, 250 x 200 pixels in 256 gray levels | 4 | 100 % |
| | 8 | 100 % |
| | 16 | 100 % |

## 6. Conclusion

This paper proposes a new method of selecting the ROI for analysis of palm print that utilizes the maximum palm region of a person suitable for feature extraction and authentication of individuals by using their palm-print features. The method uses simple edge operators to compare the 'busyness' to eliminate rectangular areas from the boundaries of the original image which may not contribute to meaningful texture feature. The algorithm can be implemented using data parallelism to reduce processing time complexity. Moreover, the ROI decreases the computations required for feature extraction. The principle used can generate dynamically adaptive ROI suiting the feature extraction method used for the identification.To verify that the palm region extraction will give better result, this region was presented to the palm print image verification mechanism that was proposed earlier [12]. The comparison results in Table 1 shows marked improvement over authentication without ROI pre-processing. An analysis of the histogram of the extracted ROI shows in fig.3 indicates a more uniform distribution of grey values, the trailing values and bimodal nature in the original picture histogram having been eliminated. The authors expect that such a ROI pre-processing can be of use to any type of Biometric processing application.

## 7. References

[1] D Zhang, W.Shu."*Two novel Characteristics in palmprint verification: datum point invariance and line feature matching"*. Pattern Recognition. 32 (4), 1999, pp. 691-702.

[2] A.K. Jain, R. Bolle, S. Pankanti, "*Biometrics: Personal Identification in Networked Society"*, Kluwer Academic Publishers, Dordrecht, 1999.

[3] Jane Youa, Wenxin Lia, David Zhanga,. *"Hierarchical palmprint identification via multiple feature extraction"*. Pattern Recognition. 35, 2002, pp. 847–859.

[4] Chin-Chuan Han, Hsu-Liang Cheng, Chih-Lung Lin , Kuo-Chin Fan "*Personal authentication using palm-print features"* Pattern Recognition 36 (2003) 371 – 381.

[5] C. Poon, D.C.M Wong, H.C.Shen, "A *new method in locating and segmenting palmprint into Region-of-Interest"*. ICPR 4, 2004, p 1051-4651.

[6] Pablo Hennings and B.V.K. Vijaya Kumar "*Palmprint Recognition Using Correlation Filter Classifiers*" IEEE 2004.

[7] J.Z. Wang, J. Li, and G. Wiederhold, "SIMPLIcity: Semantics-Sensitive Integrated Matching for Picture LIbraries,"

*IEEE Trans. Pattern Anal. Machine Intell.* vol. 23, no. 9, 2001.

[8] Ying-Han Pang, Andrew T.B.J, David N.C.L, Hiew Fu San "Palmprint Verification with Moments" Journal of WSCG, Vol.12, No.1-3, ISSN 1213-6972, WSCG'2004, February 2-6, 2003, Plzen, Czech Republic.

[9] [Tee02a] Tee Connie, Michael Goh, Andrew Teoh, and David Ngo "An automated biometric palmprint verification system" 3rd Int. Symp. On Communications & Info. Tech. (ISCIT2003), vol. 2, pp. 714-719, 2002.

[10] Kasturika B. Ray , Rachita Misra "*A New Method in Palmprint Features using Region-of-Interest*", 3rd National Conference on Recent Trends in Communications computation and Signal Processing (RTCSP), Mar 1st-2nd 2011 at Koimbator, Amrita ViswaVidyalayam University. pp. 294-297.

[11] Kasturika B. Ray , Rachita Misra "*Palmprint as a Biometric Identifier"* IJECT Vol. 2, Issue 3, Sept. 2011 ISSN : 2230-7109(Online) | ISSN : 2230-9543(Print).

[12] Rachita Misra, Kasturika B. Ray "*A Texural Approach to Palmprint Identification"* International Journal of Advanced Studies in Computer Sciences and Engineering (IJASCSE)", Vol. 1, Issue. 2, Sept -2012, ISSN: 2278-7917 (Online).

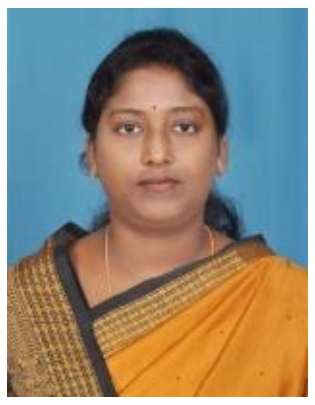

**Kasturika B. Ray** has a Post graduate degree in MIT (Master of Information Technology), Manipal Deemed University, Karnataka and continuing her Ph.D. research in the field of Digital Image processing in Computer Science and Engineering, SOA(Siksha 'O' Anusandhan) University, Bhubaneswar, India under the guidance of Dr. Mrs. Rachita Misra.

Currently she is working as a Dean Academics in Eklavya College of Engineering Technology and Science, Kusumati, Bhubaneswar, Odisha, India. She has nearly eight years of teaching experience. She has several publications in the areas of Image Processing in International and National journals, Seminars and Conferences.

She has published six International Journal research papers and presented in two National conferences and has attended ten National Workshops / Seminars etc. She has one article. Her area of interest is Digital Image Processing and Networking.

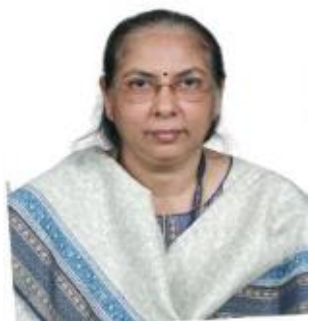

**Dr. Rachita Misra** has a Post graduate degree in Mathematics and Ph.d in the field of Digital Image processing. She has around 25 years of industrial experience in Information Technology solutions and consultancy. She has nearly 10 years of research / teaching experience. She has several publication in the areas of Image Processing, Data Mining and Software Engineering in international and national journals, seminars and conferences.
She is currently heading the Information Technology Department of C.V.Raman College of Engineering, Bhubaneswar, India. She is the editor of International Journal of Image Processing and Vision Science and technical reviewer of several international and national conferences. She is life member of Computer Society of India (CSI), Indian unit of Pattern Recognition and Artificial Intelligence (IUPRAI), Indian Science Congress Association (ISCA) and Odisha Information Technology Society (OITS).